\documentclass{article}
\usepackage{ijcai21}

\usepackage{times}
\usepackage{soul}
\usepackage{url}
\usepackage[hidelinks]{hyperref}
\usepackage[utf8]{inputenc}
\usepackage[small]{caption}
\usepackage{graphicx}
\usepackage{amsmath}
\usepackage{amsthm}
\usepackage{booktabs}
\usepackage{amssymb}
\usepackage{color}

\title{SiamRCR: Reciprocal Classification and Regression for Visual Object Tracking}

\author{
Jinlong Peng$^{1*}$
\and
Zhengkai Jiang$^{1*}$\and
Yueyang Gu$^{1}$\thanks{\textit{Equal contribution.}}\and
Yang Wu$^{2}$\thanks{\textit{Corresponding author: Yang Wu (wuyang0321@gmail.com).}}\and \\
Yabiao Wang$^1$\and
Ying Tai$^1$\and
Chengjie Wang$^1$\And
Weiyao Lin$^3$
\affiliations
$^1$Tencent Youtu Lab\\
$^2$Kyoto University\\
$^3$Shanghai Jiao Tong University
\emails
\{jeromepeng, zhengkjiang, yueyanggu, caseywang, yingtai, jasoncjwang\}@tencent.com,
wu.yang.8c@kyoto-u.ac.jp,
wylin@sjtu.edu.cn
}

\begin{document}

\maketitle

\begin{abstract}
Recently, most siamese network based trackers locate targets via object classification and bounding-box regression. Generally, they select the bounding-box with maximum classification confidence as the final prediction. This strategy may miss the right result due to the accuracy misalignment between classification and regression. In this paper, we propose a novel siamese tracking algorithm called SiamRCR, addressing this problem with a simple, light and effective solution. It builds reciprocal links between classification and regression branches, which can dynamically re-weight their losses for each positive sample. In addition, we add a localization branch to predict the localization accuracy, so that it can work as the replacement of the regression assistance link during inference. This branch makes the training and inference more consistent. Extensive experimental results demonstrate the effectiveness of SiamRCR and its superiority over the state-of-the-art competitors on GOT-10k, LaSOT, TrackingNet, OTB-2015, VOT-2018 and VOT-2019. Moreover, our SiamRCR runs at 65 FPS, far above the real-time requirement.
\end{abstract}

\section{Introduction}
As one of the fundamental research topics in computer vision, visual object tracking (VOT) plays an important role in many applications such as human-computer interaction, visual surveillance, medical image processing, and so on. It aims to locate objects in subsequent sequences according to a given ground-truth for each object target in a chosen video frame where the target appears. There is no prior knowledge about object class, which is the most significant characteristic of object tracking. Although researchers have paid much attention to object tracking, it is still a challenging task when any of the following factors exists significantly: occlusion, deformation, and scale variation.

\begin{figure}[!ht]
\centering
\includegraphics[width=\columnwidth]{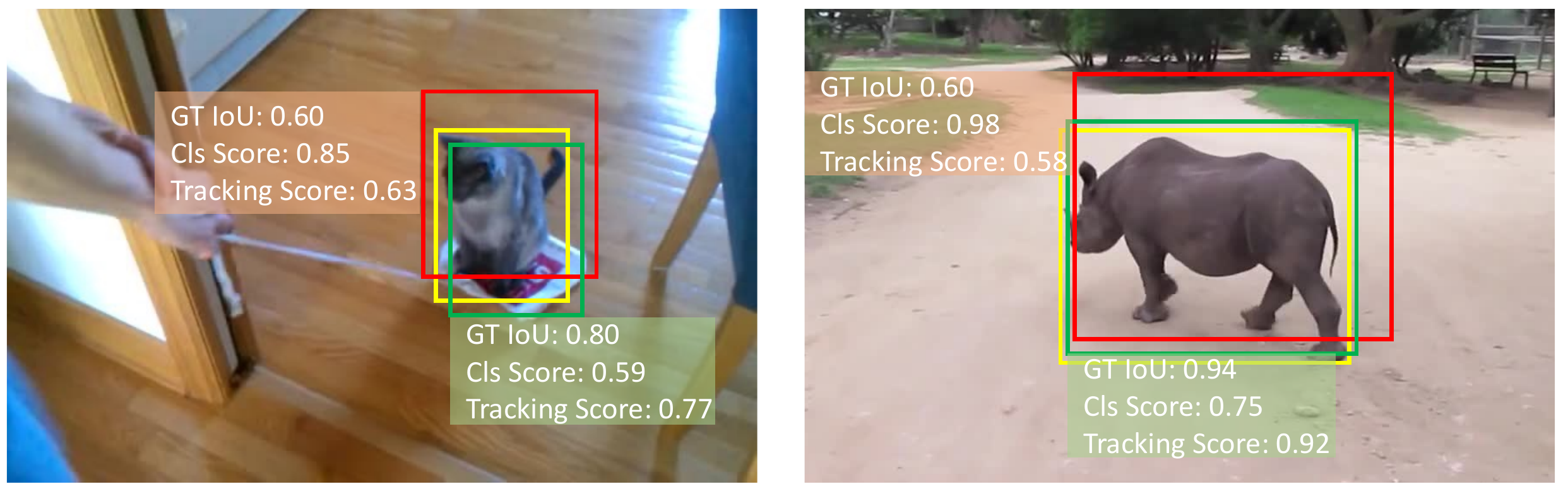}
\caption{Case study on the accuracy misalignment problem between classification and regression of siamese network based tracking models and our solution. The yellow bounding boxes denote the ground-truths, while the red and green bounding boxes are the winners ranked by the classification score and the proposed tracking score, respectively. Clearly, the tracking scores generated by SiamRCR are much more consistent with the localization/regression accuracy values (IoU), leading to better tracking performance.}
\label{cls-vs-loc}
\end{figure}

\begin{figure*}
\centering
\includegraphics[width=0.95\textwidth]{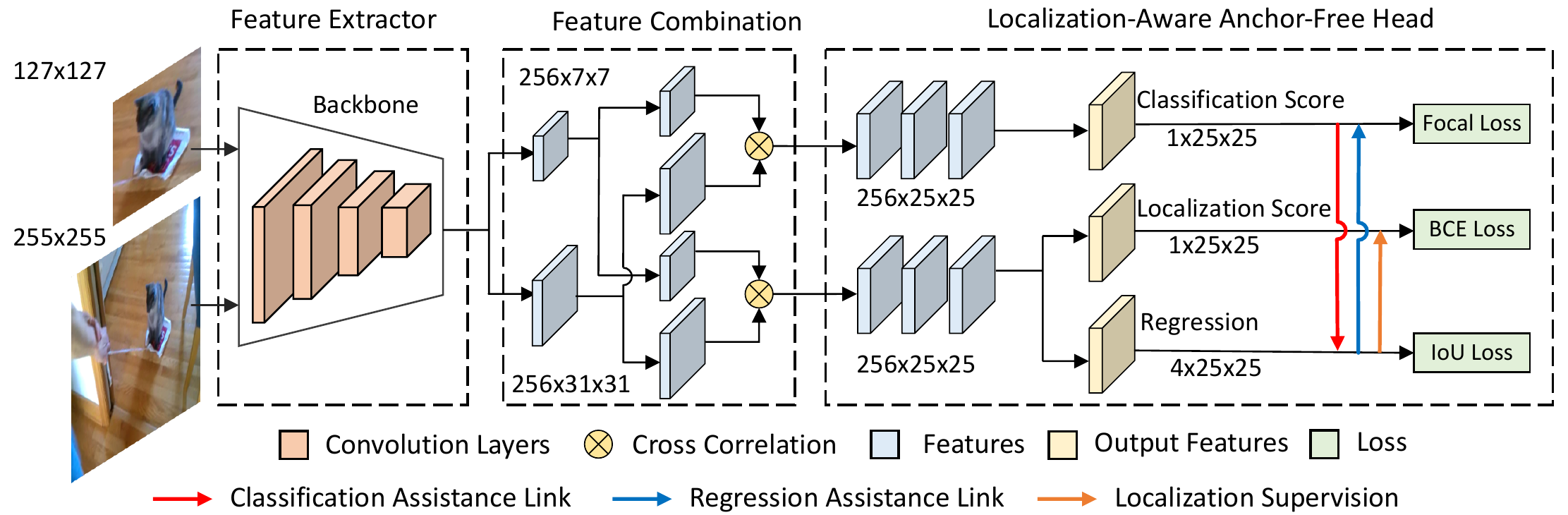}
\caption{Our proposed siamese framework on Reciprocal Classification and Regression (SiamRCR). It consists of a feature extractor, a feature combination module, and a three-branch siamese head structure with the novel reciprocal links over the individual losses. Note that the three links between the three branches are only designed for loss calculation during training and do not exist during inference.}
\label{fig:framework}
\end{figure*}

%~\cite{SiamFC,CFNet, trisiam, RASNet, dsiam, twofold_siamese, SiamRPN, SiamRPN++, SiamFC++, SiamBAN, Ocean}.
%similarity metric on lager-scale datasets off-line
% This kind of trackers formulate tracking as \ywu{a} matching problem and \ywu{thus take both the target region and the whole new video frame as their inputs}. 
Recently, siamese network based tracking has attracted increasing interest due to its balance between accuracy and efﬁciency \cite{SiamFC,SiamRPN,Ocean}. A siamese network consists of two branches sharing the same parameters for feature extracting. Exemplar image (ground-truth in the first frame) and search image (ROI of a frame to be tracked in) are inputs to the siamese network. After feature extraction and cross-correlation, it breaks into two branches: a classification branch outputs a confidence map for position estimation and a regression branch predicts the target bounding box information corresponding to each position of the confidence map. Such a network structure allows a straight-forward inference method: finding the maximum value on the 2D confidence map (from the classification branch) and then using its position to get the corresponding regressed bounding box information (from the regression branch). However, such a siamese structure generally has classification and regression optimized independently and all existing models have failed to make them properly synchronized. This results in the \textbf{accuracy misalignment between classification and regression}. As shown in Figure~\ref{cls-vs-loc}, the predicted box with high classification confidence may not have high regression accuracy in terms of IoU (Interaction over Union) score. Due to the misalignment, the bounding-box which locates the target more accurately than others might be discarded, leading to an inferior tracking performance. Although some recent siamese network \cite{ATOM,SiamFC++} have tried to predict the localization/regression accuracy, the misalignment over there is still severe since the independent optimization issue of classification and regression remains unsolved.

In this paper, we propose a novel solution to alleviate the misalignment, which builds a reciprocal relationship between classification and regression, so that they can be optimized in a synchronized way for generating accuracy consistent outputs. Since the reciprocal relationship is the key for its success, we name our model Siamese Network based Reciprocal Classification and Regression with \textbf{SiamRCR} as its abbreviation. The overall framework of SiamRCR is shown in Figure~\ref{fig:framework}. Besides the commonly used classification branch and regression branch, we add two links (the classification assistance link and the regression assistance link) to build the reciprocal relationship between them during model training. Classification assists regression by weighting the regression loss with the classification confidence, so that regression can focus more on high confident positions for more precise location. Regression assists classification by weighting the classification loss with the localization score derived from the regressed bounding box and the ground-truth box, forcing classification score to be more consistent with regression accuracy. Since there is no such localization score during testing/inference (ground-truth bounding box is unknown), a localization branch is added to predict such a localization score at each position, so that the prediction can be used as localization score's approximation to be consistent with the training model. Therefore, the multiplication of the classification confidence and the localization prediction confidence generates a new tracking score/confidence map for regression during testing, which ensures the consistency with the training process. 
%The location confidence is trained by dynamic label: IoU between predicted bounding-box and ground-truth of targets. Therefore, the location confidence can \ywu{indicate} exact localization states. Besides, it is also significant that classification assists regression. We use prediction of classification to re-weight regression loss. The predicted box with high confidence will be better optimized during training. This Reciprocal classification and regression can alleviate the misalignment effectively. 

Besides the key idea of reciprocal classification and regression, two other designs also contribute to the effectiveness and superiority of our model. One is that we choose to build on the anchor-free tracking mechanism so that the whole model can be one-stage, clean, efficient with fewer hyper-parameters. The other is that our model predicts center offset and width/height of the target, which is more straightforward and efficient than other VOT methods.

%Above-mentioned anchor-based (SiamRPN++) and anchor-free (SiamFC++) trackers locate bounding-box by predicting \emph{offsets}, which is not straightforward. Given that a bounding box can be uniquely determined by a 4-D vector (x, y, w, h), we propose an efficient siamese tracking framework, which estimates bounding-boxes of targets \emph{directly}. The framework named SiamWH contains confidence branch and box branch in order to estimate center point (x, y) and width/height (w, h) respectively. Confidence branch outputs classification confidence map and location confidence map. Classification confidence map indicates whether a pixel is positive sample or not. Location confidence map further indicates localization accuracy of positive samples. These two map estimate center point jointly. Box branch output weight map and height map, which estimate scale of the box jointly. The channel of each map is only \emph{1}. Thus, SiamWH outputs fewer predictions than SiamRPN++ or SiamFC++. We integrate the proposed localization-aware tracking method into SiamWH. 

The main contributions of this work are listed as follows:

1. We propose a novel tracking model that solves the long-standing unsolved classification and regression misalignment problem, with new simple, intuitive and efficient designs. %  and utilize location map to predict IoU.

2. It presents a new way on how to link losses of multiple branches and make the training and inference process more consistent, which may provide inspirations to other tasks. 

3. Our SiamRCR achieves state-of-the-art performance on six public benchmarks, including GOT-10k, TrackingNet, LaSOT, OTB-2015, VOT-2018 and VOT-2019. The framework is built on an anchor-free mechanism with a more direct center offset and width/height prediction, running at 65 FPS.  

%3. \ywu{\jzk{The overall framework} is built on an anchor-free mechanism with a more direct center offset and width/height prediction, so that it can be light and efficient, running at 65 FPS.}

%4. Our \pjl{SiamRCR} achieves state-of-the-art performance on six public benchmarks, including GOT-10k, TrackingNet, LaSOT, OTB-2015, VOT-2018 and VOT-2019.  

\iffalse
\begin{itemize}
    \item We propose a novel tracking model that solves the long-standing unsolved classification and regression accuracy misalignment problem, with new simple, intuitive and efficient designs. %  and utilize location map to predict IoU.
    \item It presents a new way on how to link losses of multiple branches and making the training model and inference model consistent, which may be inspiring for other tasks beyond tracking, such as object detection. 
    \item The proposal is built on an anchor-free mechanism with a more direct bounding box representation, so that it can be light and efficient, running at 65 FPS.
    \item Extensive experiments on six public benchmarks: GOT-10k~\cite{got},
    TrackingNet~\cite{trackingnet}, LaSOT~\cite{lasot}, OTB-2015~\cite{OTB2015}, VOT-2018~\cite{vot2018} and VOT-2019~\cite{vot2019} demonstrate that our SiamRCR achieves state-of-the-art performance.  
    
\end{itemize}
\fi

\section{Related Works}

\subsection{Siamese Network based Framework}
Comparing with traditional correlation filter tracking methods, recent siamese network based methods have achieved superior performance since the pioneering work SiamFC was proposed \cite{SiamFC}. More recent studies \cite{SiamRPN,SiamRPN++} try to introduce object detection progresses into object tracking for more accurate location prediction. Though these works have explored several important aspects, the accuracy misalignment problem between classification and regression has been overlooked. Ocean~\cite{Ocean} partially concerns a similar issue and presents a feature alignment module to alleviate it by utilizing the prediction of regression branch to refine the classification branch. However, this cannot eliminate the misalignment problem as the alignment is monodirectional. Differently, our SiamRCR focuses on the misalignment problem and proposes a simple, intuitive and more thorough solution with bidirectional and reciprocal links and a novel complementary branch for making training and inference consistent.

\subsection{Anchor-Free Tracking Mechanism}
%Both object detection and object tracking require to locate targets by bounding-boxes. Therefore, researchers try to introduce object detection method into object tracking. SiamRPN~\cite{SiamRPN} predicts bounding-box via RPN, which is first proposed in Faster-RCNN~\cite{faster-rcnn} for object detection. It solves tracking task as one-shot detection. RPN predicts offsets based on anchors for more accurate location prediction than SiamFC. In~\cite{SiamRPN++}, the authors improved SiamRPN by deeper backbone and random shift sampling strategy. 

Anchor-free methods have recently attracted widespread attention in the object detection field~\cite{law2018cornernet,duan2019centernet,tian2019fcos,zhou2019objects} due to their simplicity and efficiency. Naturally, the anchor-free mechanism has also been introduced to the tracking field \cite{SiamFC++,SiamBAN,Ocean}. Multiple object tracking (MOT) is a related area of VOT~\cite{peng2020dense,peng2020tpm}. In MOT area, based on CenterNet~\cite{zhou2019objects}, CenterTrack~\cite{zhou2020tracking} obtains high performance by predicting the center point, width/height and center offset of each object. To our best knowledge, SiamRCR is the first VOT method predicting center offset and width/height of the target, which is more straightforward and efficient than ever.

\subsection{Dynamic Sample Re-weighting}
Existing trackers \cite{SiamRPN,SiamRPN++,SiamFC++,peng2020ctracker} directly use some heuristic rules, \emph{e.g.}, the Focal Loss \cite{lin2017focal} to define the labels of samples and their weights. PrDiMP \cite{danelljan2020probabilistic} models the uncertainty of the labels. Such predefined static weights lead to the accuracy misalignment problem between classification and regression, which harms the final tracking accuracy. However, in our SiamRCR, the sample weights for each loss become dynamic as they are conditioned on the other branch's outputs which keep changing during the interaction. Such dynamic sample re-weighting mechanism is novel and also critical to the effectiveness of our model. 

\subsection{Localization Prediction Strategy}
In object detection area, IoU-Net \cite{jiang2018acquisition} predicts the IoU
between each detected box and the matched ground-truth to guide the box regression, which is class-specific thus not directly suitable for VOT. ATOM \cite{ATOM} trains a target-specific IoU prediction network offline and SiamFC++ \cite{SiamFC++} estimates the bounding box quality based on centerness \cite{tian2019fcos}. However, both the purpose and implementation of the localization branch in our SiamRCR are different. Our localization branch is a natural auxiliary of the reciprocal classification and regression structure which itself is a better solution than existing works, while the IoU network in other works is the main. 
%Our SiamRCR uses the multiplication of the classification score and the localization score to select bounding box instead of the predicted IoU score in ATOM. 
Moreover, our localization branch is simple and lightweight, which ensures the effectiveness and efficiency of the algorithm simultaneously.

\section{Proposed Method}

\subsection{Overview}
The proposed siamese tracking framework is shown in Figure~\ref{fig:framework}. Different from previous anchor-based~\cite{SiamRPN,SiamRPN++} methods which rely on pre-defined anchor sizes and scales, our method is anchor-free. It operates as follows. First, the target template and the current frame are both fed into the shared feature extractor (using the backbone of \cite{resnet}) to generate their corresponding features. Then, such features are combined through depth-wise cross-correlation operation to create correlated feature maps, which are further fed into the corresponding classification and regression branches of the anchor-free tracking head. The built-in reciprocal links dynamically re-weight the samples for computing each loss of the two branches. A new localization branch grows from the regression branch for predicting the localization accuracy. Its output can serve as the approximation of the localization score during inference to generate a more accurate tracking score together with the classification confidence. The key components are in detail as below.

\subsection{Anchor-Free Tracking with Box Regression}
%SiamRPN and SiamFC++ tracking targets are pre-defined anchors and offsets, respectively. Alternatively, we propose an efficient anchor-free tracking called SiamWH (siamese tracking via width/height and center offsets). SiamWH contains two branch: confidence branch and box branch. Confidence branch outputs classification confidence map $1\times25\times25$. And box branch outputs regression map $4\times25\times25$ and the corresponding localization map $1\times25\times25$ following~\cite{tian2019fcos}.
For the $i$-th input pair from the training set, we have $F_i \in \mathcal{R}^{C \times H \times W}$ denotes the feature map of the classification branch and $s$ be the total stride. The ground-truth bounding box for the current frame is defined as $B_{x,y}^* = (x_0^*, y_0^*, x_1^*, y_1^*)$, \emph{i.e.}, coordinates of the bounding box. For each location $(x, y)$ on the feature map $F_i$, we can map it back onto the input frame to get the corresponding image coordinates $(\lfloor \frac{s}{2} \rfloor + xs,\lfloor \frac{s}{2} \rfloor + ys)$. Different from anchor-based trackers, which consider the location on the input frame as the center of anchor boxes and regress the target bounding boxes w.r.t. the anchor boxes, we directly regress the target boxes' width and height values and the center offsets at the location. In this way, our tracker views locations as training samples instead of anchor boxes, which follows the paradigm of the FCNs~\cite{long2015fully} for semantic segmentation.

Specially, the sample at location $(x, y)$ is considered to be positive if it falls into a radius $r$ at the ground-truth box center, and the radius is a hyper-parameter for the proposed method. Otherwise, it is a negative sample (background). Besides the label (denoted by $c_{x,y}^*$) for foreground-background classification, we also have a 4D real vector $t_{x,y}^*=(w^*, h^*, \Delta x^*, \Delta y^*)$ indicating the regression target for the localization. Here, $w^*$ and $h^*$ are the width and height of target ground-truth bounding box, while $\Delta x^*$ and $\Delta y^*$ are the center offsets between the current location and the ground-truth box. Formally, if location $(x, y)$ is associated to the ground-truth box $B_{x,y}^*$, which has width $w^*$ and height $h^*$, then we have%the training regression targets for the positive location can be formulated as follows:
\begin{equation} \label{Eq1}
\setlength{\abovedisplayskip}{6pt}
\setlength{\belowdisplayskip}{6pt}
\begin{split}
w^* = x_1^* - x_0^*&, \quad h^* = y_1^* - y_0^*,  \\
\Delta x^* = (x_0^* + x_1^*) / 2 - x&, \quad \Delta y^* = (y_0^* + y_1^*) / 2 - y.  
% \vspace{-1em}
\end{split}
%\vspace{-4em}
\end{equation}

Corresponding to the training target, SiamRCR predicts a classification confidence score $p_{x,y}^{cls}$, a regressed 4D vector $t_{x,y}=(w, h, \Delta x, \Delta y)$ for the bounding box, and a localization confidence score $p_{x,y}^{loc}$ denoting the predicted localization accuracy. It is worth noting that SiamRCR has $5 \times$ fewer network parameters than the popular anchor-based tracker SiamRPN \cite{SiamRPN} with 5 anchor boxes per location.

\subsection{Reciprocal Classification and Regression}

In existing siamese network tracking models, classification and regression branches operate in parallel and get optimized independently with their own losses, which aggravates the accuracy misalignment of their results. In fact, when a regressed bounding box has low accuracy, the corresponding classification score should not be high, because if that position becomes the winner of classification confidence the bad localization will lead to bad tracking performance. And when a bounding box has a low classification score, there is no meaning for the regression to try hard to get a high localization accuracy for it will not be the winner anyway. Therefore, these two branches need to talk to each other for aligning the accuracy of their results. In this paper, we propose a novel strategy called reciprocal classification and regression to make these two branches assist each other. It is implemented by building two links, including regression assistance link and classification assistance link.

%Suppose that the target center in final response map (e.g. classification confidence map) is $(\mathop{x_c}\limits^{-}, \mathop{y_c}\limits^{-})$, the positive and negative samples in response maps are divided by same strategy:
%\begin{itemize}
%    \item If the distance between a pixel and $(\mathop{x_c}\limits^{-}, \mathop{y_c}\limits^{-})$ is smaller than $r$, it belongs to positive sample.
%    \item If the distance between a pixel and $(\mathop{x_c}\limits^{-}, \mathop{y_c}\limits^{-})$ is greater than $r$, it belongs to negative sample.
%\end{itemize}

% \begin{figure}[!t]
% \centering
% \includegraphics[width=0.95\columnwidth]{./fig/centerness-vs-iou.pdf}
% \caption{The \ywu{confidence map} comparison \ywu{between the proposed} \peng{localization branch} \ywu{and} centerness~\protect\cite{tian2019fcos}. For centerness, center location has larger value, which is not suitable for most of ground-truth boxes. Our proposed target is the IoU between regressed box with ground-truth box, which dynamically changes to match localization accuracy of regressed boxes during training.}
% \label{centerness-vs-iou}
% \end{figure}

\begin{figure}[!t]
\centering
\includegraphics[width=\columnwidth]{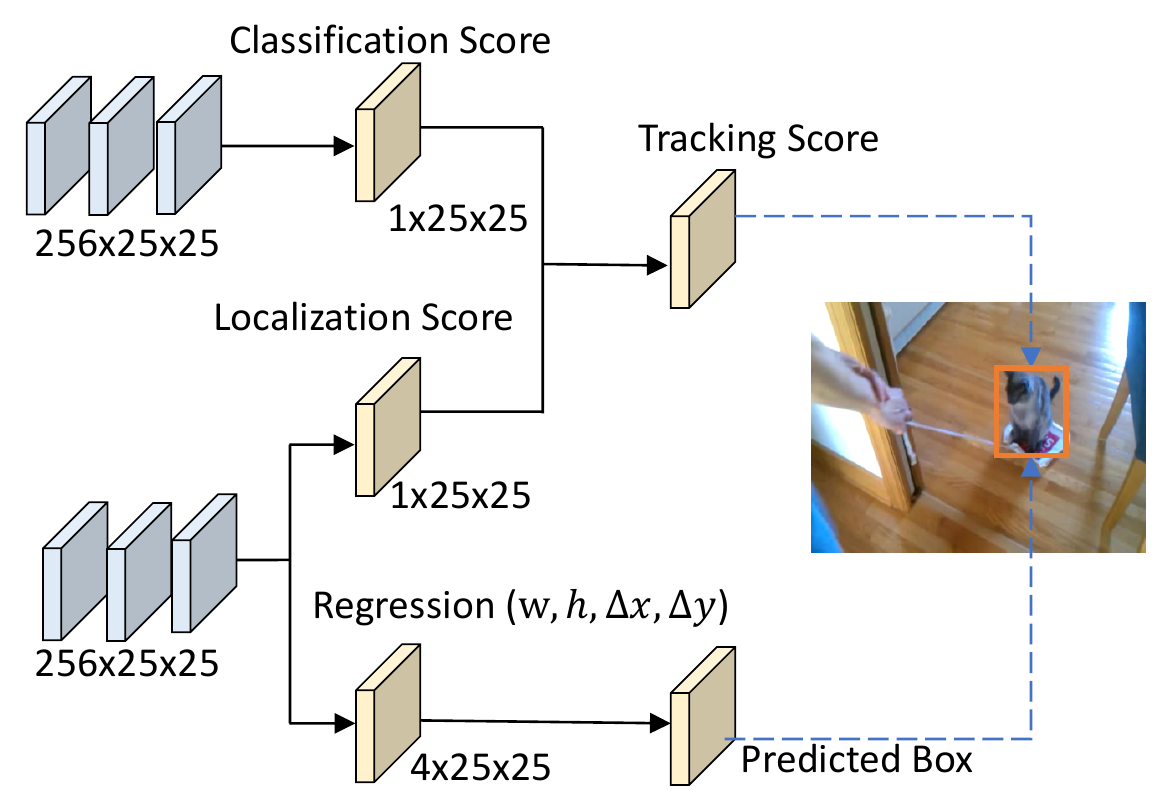}
\caption{The head of SiamRCR during inference. The classification score and localization score are multiplied to generate the final tracking score for ranking the predicted bounding boxes.}
\label{fig:wh}
\end{figure}

%\noindent \textbf{Regression Assistance Link}
\subsubsection{Regression Assistance Link}

To eliminate the chance that low localization accuracy bounding boxes still get high classification scores, a simple yet effective solution is to use the localization accuracy to weight the classification loss. Such an assistance link from regression can be regarded as a kind of dynamic sample re-weighting as the localization accuracy keeps changing during the model optimization. The dynamically re-weighted classification loss can be formulated as:
\begin{equation}
    L_{cls} = \frac{1}{N_{pos}}\sum_{x, y} L_{Focal}(p^{cls}_{x, y}, c^*_{x, y}) * IoU(B_{x, y}, B^*_{x, y}),
\end{equation}
where $L_{Focal}$ and $IoU$ denote the focal loss \cite{lin2017focal} and the IoU score, respectively, $N_{pos}$ is the number of positive samples, and $B = (x_0, y_0, x_1, y_1)$ is the predicted bounding box at location $(x, y)$ with predicted width/height $(w, h)$ and center offsets $(\Delta x, \Delta y)$:
\begin{equation} \label{Eq2}
\setlength{\abovedisplayskip}{6pt}
\setlength{\belowdisplayskip}{6pt}
\begin{split}
x_0 = x + \Delta x& - w / 2, \quad y_0 = y + \Delta y - h / 2,  \\
x_1 = x + \Delta x& + w / 2, \quad y_1 = y + \Delta y + h / 2.  
% \vspace{-1em}
\end{split}
%\vspace{-4em}
\end{equation}

%Classification confidence map is used to distinguish foreground object and background. We utilize BCE loss for this binary classification:

%\begin{equation}
%    l^{cls} = \frac{1}{N_{pos}}\sum\limits_{(x, y)\in P \cup N} l^{BCE} (Y^{cls}(x, y), R^{cls}_{x, y}) 
%\end{equation}

%where $P$ and $N$ denotes positive and negative sample sets respectively,  $R^{Cls}_{x, y}$ denotes response in location confidence map at position $(x, y)$, $Y^{Loc}_{x, y} \in {1, 0}$ denotes corresponding label. 

%Location confidence map is used to estimate localization accuracy. We expect that the location confidence map can predict the IoU between ground-truth and corresponding bounding-box. Thus we calculate the IoU of bounding-boxes from box branch and utilize it as dynamic label for training location map. Here, we only take the localization loss of positive samples into account:
%\begin{equation}
%    l^{loc} = \frac{1}{N_{pos}}\sum\limits_{(x, y)\in P} l^{BCE} (IoU(B^*, B_{x, y}), R^{Loc}_{x, y}) 
%\end{equation}
%where $B^*$ denotes ground-truth of targets, $B_{x, y}$ denotes predicted bounding-box at position $(x, y)$, $R^{Loc}_{x, y}$ denotes response in location confidence map at position $(x, y)$.

%\noindent \textbf{Classification Assistance Link} 
\subsubsection{Classification Assistance Link}

To avoid low confidence positions getting highly accurate bounding boxes, the regression branch should be aware of the classification confidence. To this end, $p_{x, y}^{cls}$ is utilized to dynamically re-weight the regression loss as: 
\begin{equation}
    L_{reg} = \frac{1}{N_{pos}}\sum_{{x, y}}{\mathbb{I}_{\{c^*_{x, y} = 1\}}L_{IoU}(t_{x, y}, t^*_{x, y})} * p_{x, y}^{cls},
\end{equation}
where $L_{IoU}$ is the IoU loss as in UnitBox~\cite{yu2016unitbox}; $\mathbb{I}_{\{c^*_{x, y} = 1\}}$ is an indicator function, which equals to $1$ if $c^*_{x, y} = 1$ and $0$ otherwise.

%We also utilize confidence classification to assist bounding-box regression. Since we select final bounding-box according to classification confidence, the regression loss of predicted bounding-box with high confidence is more significant. Therefore, we use classification to re-weight IoU loss:

%\begin{equation}
%     l^{reg} = \frac{1}{||P||} \sum\limits_{(x, y)\in P} R^{Cls}_{x, y} \cdot IoULOSS(B^*, B_{x, y})
%\end{equation}

%where $R^{Cls}_{x, y}$ denotes response in classification confidence map at position $(x, y)$. This regression loss can focus on predicted box with high classification confidence benefit from re-weighting. 

%The overall loss is:

%\begin{equation}
%    l = l^{cls} + \lambda_1 l^{loc} + \lambda_2 l^{reg}  
%\end{equation}
%where $\lambda_1$ and $\lambda_2$ denotes weights for localization confidence loss and regression loss respectively.  

%\subsection{IoU-Aware \ywu{Inference}}

%\noindent \textbf{Localization Score Branch} 
\subsubsection{Localization Score Branch}

%After using the proposed anchor-free framework, there is a still performance gap between our method and anchor-based detectors. We observed that it is due to a lot of low-quality predicted bounding boxes. We further propose a simple yet effective strategy to suppress the low-quality bounding boxes with high classification scores. Specifically, we add an IoU-aware single branch, in parallel with the regression branch (as shown in \ywu{Figure}~\ref{fig:framework}). 

The regression assistance link makes the classification
branch aware of the regression accuracy during training, thanks to the ground-truth bounding box $B_{x, y}^*$ for computing the localization score. However, in the inference stage there is no such ground-truth. Directly using the classification confidence map $p^{cls}$ to select the winner bounding box may still lead to certain accuracy misalignment, as the localization score was hands-on during the classification branch's training. The hands-on inductive training makes $p^{cls}$ collaborative with the localization score but not necessarily consistent with it. Therefore, we let the regression branch grow a new branch called localization branch to be trained for predicting the localization score given the feature maps for regression, under the following loss function.
\begin{equation}
\label{iou-loss}
    L_{loc} = \frac{1}{N_{pos}}\sum_{x, y} \mathbb{I}_{\{c^*_{x, y} = 1\}}L_{BCE}(p^{loc}_{x,y}, IoU(B_{x, y}, B^*_{x, y})),
\end{equation}
where $L_{BCE}$ is the Binary Cross Entropy (BCE) loss. 
%As shown in Figure~\ref{centerness-vs-iou}, IoU between regressed boxes with ground-truth boxes could dynamically match localization accuracy during training compared with Centerness ~\cite{tian2019fcos}.

As shown in Figure~\ref{fig:wh}, during inference, the final tracking score (used for ranking the predicted bounding boxes) is computed by multiplying $p^{cls}_{x,y}$ with $p^{loc}_{x,y}$, making the inference localization-aware. Thus, the localization branch can further reduce the low-quality boxes and improve the overall tracking accuracy.

%\noindent \textbf{The Overall Training Objective}
\subsubsection{The Overall Training Objective}

With the above losses for SiamRCR's three branches, we can define its final training loss function as:
\begin{equation} 
\label{eq:loss_function_final}
L = L_{cls} + \lambda_{1} * L_{reg} + \lambda_{2} * L_{loc}.
\end{equation}
where $\lambda_{1}$ and $\lambda_{2}$ are the hyper-parameters for balancing these losses. In our experiments, they are all set to $1$.

% \begin{equation} 
% \label{eq:loss_function_final}
% \begin{split}
% L &= L_{cls} + \lambda_{1} * L_{reg} + \lambda_{2} * L_{iou} \\
% &= \frac{1}{N_{pos}} \sum_{x, y}{L_{cls}(p_{x, y}, c^*_{x, y})} * IoU(B_{x, y}, B_{x, y}^*) \\
% & \quad + \frac{\lambda_{1}}{N_{pos}}\sum_{x, y}{\mathbbm{1}_{\{c^*_{x, y} = 1\}}L_{reg}(t_{x, y}, t^*_{x, y})} * p_{x, y} \\
% & \quad + \frac{\lambda_{2}}{N_{pos}}\sum_{x, y}{\mathbbm{1}_{\{c^*_{x, y} = 1\}}L_{iou}(IoU_{x, y}, IoU^*_{x, y})},
% \end{split}
% \end{equation}

%where $L_{cls}$ is focal loss as in \cite{lin2017focal}, $L_{reg}$ is the IoU loss as in UnitBox~\cite{yu2016unitbox} and $L_{iou}$ is BCE loss. $N_{pos}$ denotes the number of positive samples, $\lambda_{1}$ and $\lambda_{2}$  being $1$ in this paper are the balance weight for $L_{reg}$ and $L_{iou}$. The summation is calculated  over all locations on the feature maps $F_i$. $\mathbbm{1}_{\{c^*_i = 1\}}$ is the indicator function, being $1$ if $c^*_i = 1$ and $0$ otherwise.

\section{Experiments}
\subsection{Implementation Details}

\paragraph{Training Phase.} We utilize ResNet-50~\cite{resnet} as the backbone of our SiamRCR. We remove the last conv-block for higher resolution feature map and utilize dilated convolution for higher receptive field~\cite{SiamRPN++}. The backbone is initialized with the parameters pre-trained on ImageNet~\cite{imagenet}. The whole network is optimized by Stochastic Gradient Descent (SGD) with momentum 0.9 on the datasets of GOT-10k~\cite{got}, TrackinegNet~\cite{trackingnet}, COCO~\cite{lin2014microsoft}, LaSOT~\cite{lasot}, ImageNet VID~\cite{imagenet} and ImageNet DET~\cite{imagenet}. We totally train the network for 20 epochs. The batch size is 128. The learning rate is from 0.000001 to 0.1 in the first 5 epochs for warm-up and from 0.1 to 0.0001 with cosine schedule in the last 15 epochs. We freeze the backbone in the first 10 epochs and fine-tune it in the other 10 epochs with a reduced learning rate (multiplying $0.1$). The size of exemplar image and search image are 127*127 and 255*255, respectively. Our algorithm is implemented by Python 3.6 and PyTorch 1.1.0. The experiments are conducted on a server with Intel(R) Xeon(R) CPU E5-2680 v4 2.40GHz, and a NVIDIA Tesla P40 24GB GPU with CUDA 10.1.

\paragraph{Testing Phase.} We utilize the same offline testing strategy as~\cite{SiamFC++}. The ground-truth after augmentation in the first frame is used as the exemplar image and we keep it unchanged during the whole testing phase. A cosine-window~\cite{SiamFC} is multiplied on the confidence map. We adopt a linear interpolation updating strategy on scale prediction to make the final box change smoothly over time. We evaluate SiamRCR on six public benchmarks following their corresponding protocols: GOT-10k~\cite{got},
    TrackingNet~\cite{trackingnet}, LaSOT~\cite{lasot}, OTB-2015~\cite{OTB2015}, VOT-2018~\cite{vot2018} and VOT-2019~\cite{vot2019}.

\begin{figure}[t]
\centering
\includegraphics[width=0.95\columnwidth]{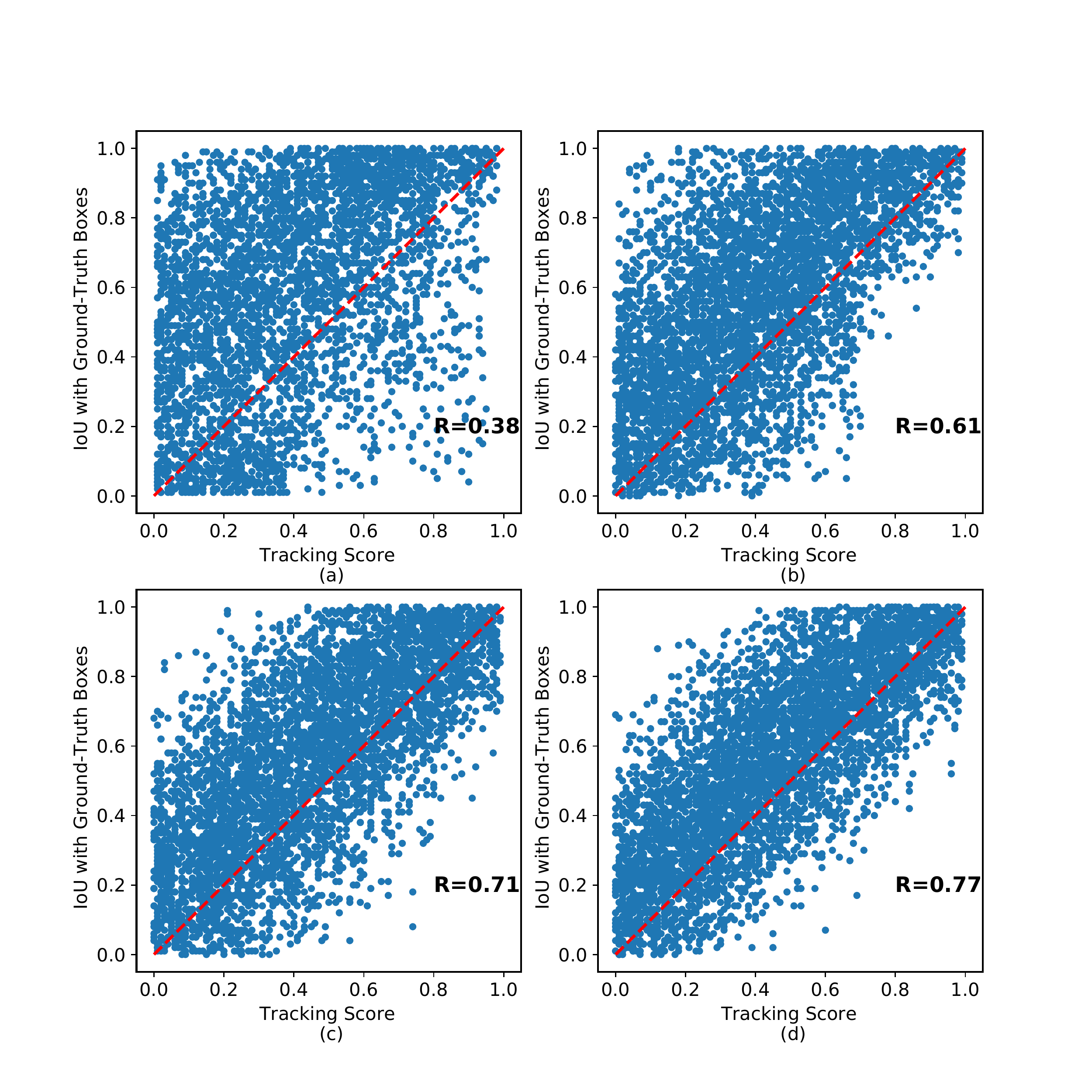}
\caption{The correlation between IoU scores and the tracking score, together with the Pearson correlation coefficient $R$. (a) Baseline model, the tracking score is the classification score. (b) Using centerness proposed in FCOS~\protect\cite{tian2019fcos} as the tracking score. (c) Baseline + localization branch. (d) SiamRCR. 
}
\label{fig:ablation-studies}
\end{figure}

\subsection{Ablation Study}

\paragraph{Component.} The ablation study results on the key components of SiamRCR are presented in Table~\ref{fig:component}. The baseline (\uppercase\expandafter{\romannumeral1}) without localization branch and reciprocal links obtains an AO (Average Overlap) of 0.594. With localization branch, SiamRCR can predict the localization score of the regressed bounding box, making the final tracking score more consistent with the real IoU than the classification score. Multiplying the localization score alone (\uppercase\expandafter{\romannumeral2}) improves the performance by 3.54\% compared with baseline, showing the significance of the accuracy misalignment between classification and regression. Building reciprocal assistance links itself (\uppercase\expandafter{\romannumeral3}) can also gain a relative improvement of 2.86\% over the baseline, proving that the misalignment can be alleviated between classification and regression. When these two components are both adopted, the relative performance is more remarkable: 5.05\%, which is nearly equal to the direct sum of both performance gains. It confirms that the localization branch is consistent with the reciprocal links, serving well as the replacement of the regression assistance link for inference. To better demonstrate how well our SiamRCR alleviates the accuracy misalignment problem, we illustrate the correlation between the IoU of regressed bounding box (w.r.t. the matched ground-truth) and the tracking score in Figure~\ref{fig:ablation-studies}. As shown in Figure~\ref{fig:ablation-studies}(a), the Pearson correlation coefficients between IoU and tracking score is only 0.38, showing that the classification score is indeed not consistent with the real localization accuracy. Figure~\ref{fig:ablation-studies}(c) and~\ref{fig:ablation-studies}(d) show that both the localization branch and the reciprocal links are effective and necessary, and they can well collaborate with each other.

%Figure~\ref{fig:ablation-studies}(b) and Figure~\ref{fig:ablation-studies}(c) shows that other solutions like the centerness branch in FCOS~\cite{tian2019fcos} is not as good as our props osed \peng{localization branch}. This is because our \peng{localization branch} directly predicts IoU while centerness is an indirect measure based on heuristics. 

%(d) Last, we propose the reciprocal classification and regression by enduing classification loss with regression localization and regression loss with classification scores, which further improve the consistence between classifications scores and localization accuracy. \pjl{The Pearson correlation coefficients between IoU and localization score is further increased to 0.77.}

\paragraph{Predicted IoU vs. Centerness.} Centerness is pre-defined label which indicates the distance between candidates and target center. Some object detection~\cite{tian2019fcos} or object tracking~\cite{SiamFC++} algorithms utilize centerness to assist localization. In our SiamRCR, we discard this kind of fixed prior and utilize predicted IoU as dynamic supervised localization information. Thus, our localization branch can estimate the localization confidence more accurately. As shown in Figure~\ref{fig:ablation-studies} (b) and (c), our localization prediction mechanism alleviates the misalignment between classification and regression, which is better than centerness.

\begin{table}[t]
\centering
\begin{tabular}{c|cc|c}
\toprule[2pt]
& Localization Branch & Reciprocal Links & AO$\uparrow$ \\ \midrule[1pt]
\uppercase\expandafter{\romannumeral1} &              &                 &   0.594 \\
\uppercase\expandafter{\romannumeral2} & $\surd$          &                 &    0.615\\
\uppercase\expandafter{\romannumeral3} &              & $\surd$             &    0.611 \\
\uppercase\expandafter{\romannumeral4} & $\surd$          & $\surd$           &   \textbf{0.624} \\
\bottomrule[2pt]
\end{tabular}
\vspace{-0.1cm}
\caption{Ablation study on GOT-10k test set.}
\vspace{-0.1cm}
\label{fig:component}
\end{table}

\begin{table}[t]
\small
\centering
\begin{tabular}{ccccc}
\toprule[2pt]
$r$ & $R$ & AO$\uparrow$    & SR@0.5$\uparrow$ & SR@0.75$\uparrow$ \\ \midrule[1pt]
1  & 8    & 0.593 & 0.723 & 0.458  \\
2  & 16    & \textbf{0.624} & \textbf{0.752} & 0.460  \\
3  & 24    & 0.619 & 0.747 & 0.459  \\
4  & 32    & 0.612 & 0.743 & 0.446  \\
5  & 40    & 0.611 & 0.740 & \textbf{0.474}  \\
\bottomrule[2pt]
\end{tabular}
\vspace{-0.1cm}
\caption{Comparative experiment in terms of $r$ on GOT-10k test set.}
\vspace{-0.15cm}
\label{fig:radius}
\end{table}

\paragraph{Radius.} Radius $r$ is a significant hyper-parameter in our proposed anchor-free framework. It decides the division of positive samples and negative samples during training. We conduct comparative experiment in terms of $r$. The results are shown in Table~\ref{fig:radius}. $R$ is the corresponding radius of $r$ in the original input video frame, which is 8 times $r$. When $r=1$, the performance on GOT-10k is poor since the number of positive samples is too small. When $r=2$, our SiamRCR achieves the best performance. When $r=4$ or $r=5$, the positive samples are redundant since some candidates far from target center are divided into positive samples. Therefore, the performance drops compared with $r=2$ or $r=3$.

\begin{figure}[tb]
\centering
\includegraphics[width=0.92\columnwidth]{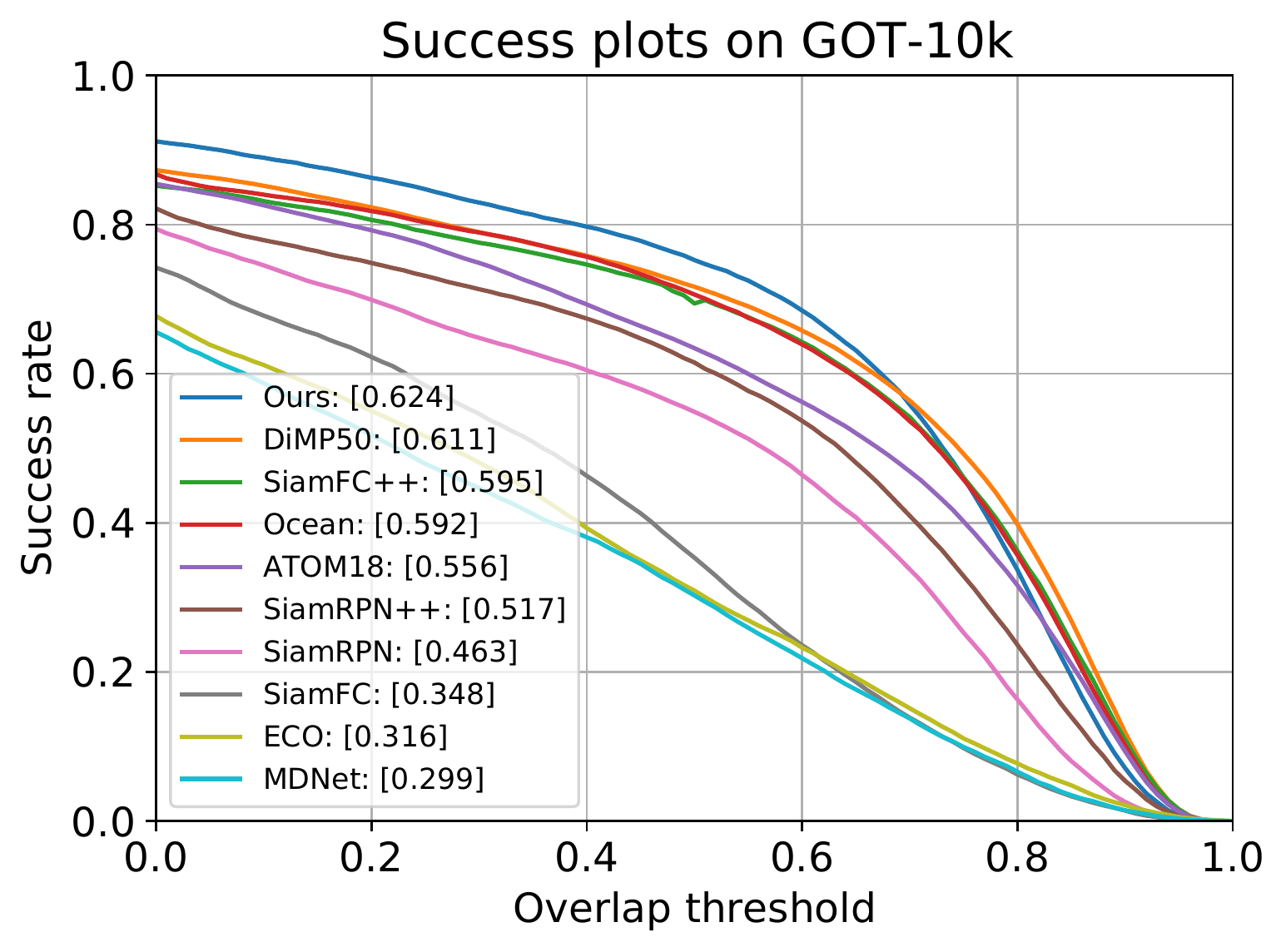}
\vspace{-0.15cm}
\caption{Comparison of tracking results on GOT-10k benchmark.}
\label{fig:got}
\end{figure}

\begin{table}[t]
\small
\centering
\begin{tabular}{cccc}
\toprule[2pt]
          & Succ.$\uparrow$ & Prec.$\uparrow$  & N-Prec.$\uparrow$ \\ \midrule[1pt]
SiamFC~\cite{SiamFC}    & 0.559 & 0.518  & 0.652  \\
ECO~\cite{ECO}       & 0.554 & 0.492  & 0.618  \\
UPDT~\cite{updatenet}      & 0.611 & 0.557  & 0.702  \\
ATOM~\cite{ATOM}      & 0.703 & 0.648 & 0.771  \\
SiamRPN++~\cite{SiamRPN++} & 0.733 & 0.694  & 0.800  \\
DiMP50~\cite{DiMP}    & 0.740 & 0.687  & 0.801  \\
KYS~\cite{KYS}    & 0.740 & 0.688  & 0.800  \\
SiamAttn~\cite{SiamAttn}  & 0.752 & \textbf{\textcolor{blue}{0.715}}  & \textbf{\textcolor{blue}{0.817}}  \\
SiamFC++~\cite{SiamFC++}  & \textbf{\textcolor{blue}{0.754}} & 0.705  & 0.800  \\
\midrule[1pt]
SiamRCR (ours)      & \textbf{\textcolor{red}{0.764}} & \textbf{\textcolor{red}{0.716}}  & \textbf{\textcolor{red}{0.818}}  \\ 
\bottomrule[2pt]
\end{tabular}
\caption{Comparison of tracking results on TrackingNet benchmark. \textbf{\textcolor{red}{Red}} and \textbf{\textcolor{blue}{blue}} fonts indicate the best and second results respectively.}
\vspace{-0.11cm}
\label{fig:trackingnet}
\end{table}

\begin{table}[t]
\small
\centering
\begin{tabular}{ccc}
\toprule[2pt]
          & Succ.$\uparrow$ & Prec.$\uparrow$ \\ \midrule[1pt]
SiamFC~\cite{SiamFC}    & 0.339 & 0.336 \\ 
MDNet~\cite{MDNet}     & 0.373 & 0.397 \\
GradNet~\cite{gradnet}   & 0.351 & 0.365 \\
SiamRPN++~\cite{SiamRPN++} & 0.491 & 0.496 \\
ATOM~\cite{ATOM}      & 0.505 & 0.514 \\
DiMP50~\cite{DiMP}    & \textbf{\textcolor{blue}{0.564}} & \textbf{\textcolor{blue}{0.568}} \\
ROAM++~\cite{ROAM}   & 0.447 & 0.445 \\
SiamBAN~\cite{SiamBAN}   & 0.514 & 0.518 \\
Ocean~\cite{Ocean}     & 0.526 & 0.526 \\
SiamFC++~\cite{SiamFC++}  & 0.544 & 0.547 \\ \midrule[1pt]
SiamRCR (ours)      & \textbf{\textcolor{red}{0.575}} & \textbf{\textcolor{red}{0.599}} \\ 
\bottomrule[2pt]
\end{tabular}
\caption{Comparison of tracking results on LaSOT benchmark.}
\vspace{0.11cm}
\label{fig:lasot}
\end{table}

\subsection{Comparison with the State-of-the-Art}
%We compare our SiamRCR with 12 state-of-the-art trackers: 2 anchor-based siamese trackers (SiamRPN++~\cite{SiamRPN++}, SiamAttn~\cite{SiamAttn}), 2 anchor-free based siamese trackers (SiamBAN~\cite{SiamBAN}, Ocean~\cite{Ocean}), 2 localization-aware trackers (ATOM~\cite{ATOM}, SiamFC++~\cite{SiamFC++}), 2 correlation filter based trackers (ECO~\cite{ECO}, DiMP~\cite{DiMP}), 2 online-updating trackers (UPDT~\cite{updatenet}, GradNet~\cite{gradnet}), 1  multi-domain learning based tracker (MDNet~\cite{MDNet}) and the original SiamFC~\cite{SiamFC}. \peng{The datasets and experimental settings are detailed as below. Due to space limitations, the experiments on OTB-2015 and VOT-2018 are displayed in supplementary.}

We compare our SiamRCR with 18 state-of-the-art trackers. The datasets and experimental settings are detailed as below. Due to space limitations, the experiments on OTB-2015 and VOT-2018 are presented in the supplementary.

\paragraph{GOT-10k.} The evaluation follows the protocols in~\cite{got}. For a fair comparison, we train SiamRCR only on the \emph{train} subset which consists of about 10,000 sequences and test it on the \emph{test} subset of 180 sequences. As shown in Figure~\ref{fig:got}, our SiamRCR achieves 0.624 of AO, which is the best among evaluated trackers (including the online updating tracker DiMP). The slightly inferior performance at large overlap threshold might due to SiamRCR's strategy of predicting the center offsets and width/height, rather than predicting the bounding box coordinate offsets (e.g. SiamFC++), as larger value ranges can lead to less preciseness. However, our strategy better solves the misalignment problem.

%These significant results prove the effectiveness of our SiamRCR for more accurate target \peng{tracking}. 

 %GOT-10k is a category-agnostic tracking benchmark, which can evaluate the generic tracking performance since there is no class intersection between \emph{train} subset and \emph{test} subset. The main meric in GOT-10k is Average Overlap (AO). 

\paragraph{TrackingNet.} The \emph{test} subset of it contains 511 sequences and 70 object classes. We also train our model only on TrackingNet \emph{train} subset. There are three metrics in TrackingNet: Success (Succ.), Precision (Prec.) and Normalized Precision (N-Prec.). We report the results in Table~\ref{fig:trackingnet}. SiamRCR surpasses other state-of-the-art trackers on all three evaluation metrics. In particular, SiamRCR obtains 0.764 of Succ., 0.716 of Prec. and 0.818 of N-Prec., which further demonstrates the superior tracking performance of our SiamRCR.
%which outperforms SiamRPN++ by of 4.3\%, 3.2\% and 2.3\%, respectively.

%The evaluation is implemented on the corresponding server refered in~\cite{trackingnet}.N-Prec. is normalized precision over the size of the ground-truth bounding box. 

\paragraph{LaSOT.} LaSOT is a large-scale long-term tracking benchmark. It contains 1,400 sequences and more than 3.5 million frames. We train our model only on LaSOT \emph{train} subset and conduct evaluation following the protocol \uppercase\expandafter{\romannumeral2} in~\cite{lasot}. As shown in Table~\ref{fig:lasot}, our SiamRCR achieves 0.575 of Succ. and 0.599 of Prec., and outperforms recent SOTA tracker Ocean by 8.5\% and 13.9\% in terms of both Success and Precision score respectively. It also achieves better performance compared with other localization-aware trackers (ATOM and SiamFC++), proving that our reciprocal links with localization branch is better.

%There are total 20 object classes and each class includes 70 sequences. Sequences in LaSOT are selected from the wild where the appearance of targets may vary considerably. 

\begin{table}[t]
\small
\centering
\setlength{\tabcolsep}{0.8mm}{
\begin{tabular}{cccc}
\toprule[2pt]
              & EAO$\uparrow$   & Accuracy$\uparrow$ & Robustness$\downarrow$ \\ \midrule[1pt]
%ATOM ~\cite{ATOM}          & 0.292 & \textbf{\textcolor{red}{0.603}}    & 0.411      \\
SPM~\cite{SPM}            & 0.275 & 0.577    & 0.507      \\
SiamRPN++~\cite{SiamRPN++}      & 0.285 & \textbf{\textcolor{blue}{0.599}}    & 0.482      \\
SiamMask~\cite{wang2019fast}      & 0.287 & 0.594    & 0.461      \\
SiamBAN~\cite{SiamBAN}       & 0.327 & \textbf{\textcolor{red}{0.602}}    & 0.396      \\
%DiMP50 ~\cite{DiMP}        & 0.321 & 0.582    & \textbf{\textcolor{red}{0.371}}      \\
Ocean~\cite{Ocean} & \textbf{\textcolor{blue}{0.327}} & 0.590    & \textbf{\textcolor{red}{0.376}}      \\ \midrule[1pt]
SiamRCR (ours)           & \textbf{\textcolor{red}{0.336}} &  \textbf{\textcolor{red}{0.602}}        & \textbf{\textcolor{blue}{0.386}}          \\
\bottomrule[2pt]
\end{tabular}}
\caption{Comparison of tracking results on VOT-2019 benchmark.}
\label{fig:vot2019}
%\vspace{0.3cm}
\end{table}

%\noindent \textbf{OTB-2015} OTB-2015 contains 98 sequences and 100 objects. In a quarter sequences of OTB-2015, frames are gray-scale images. Generally, Succ. and Prec. are used as metrics in OTB-2015. Table~\ref{fig:otb} presents the performance of evaluated trackers. Succ. score is the more reasonable metric than Prec. score. since it is normalized by the scale of boxes. Our SiamRCR outperforms SiamBAN and SiamRPN++ slightly and obtains the best Succ. score.
%As one of the most commonly-used SOT benchmark, Object Tracking Benchmark (OTB) is the first benchmark which provides fairly platform for tracking performance evaluation. 

\paragraph{VOT-2019.} With challenging factors such as occlusion, fast motion and illumination changing in 60 test sequences, VOT-2019 provides a comprehensive evaluation platform for VOT. Commonly used metrics for it are Expected Average Overlap (EAO), Accuracy and Robustness. EAO takes both Accuracy and Robustness into account to verify the overall tracking performance. We report experimental results on VOT-2019 in Table~\ref{fig:vot2019}. Our SiamRCR achieves the best EAO score, the best Accuracy score and the second best Robustness score. Ocean performs slightly better in Robustness with the multi-feature combination strategy. As our SiamRCR only uses single conv-feature for estimation, it is faster than Ocean. Moreover, it demonstrates superior effectiveness and efficiency. 

%Both VOT-2018~\cite{vot2018} and VOT-2019~\cite{vot2019} contains 60 sequences. The difference between them is that VOT-2019 replaces 10 least challenging sequences by other 10 sequences. VOT benchmark utilizes re-initialization mechanism. If a tracker fails to track targets up to 5 frames, it will be initialized with the ground-truth.

\section{Conclusion}
In this paper, we have proposed a novel anchor-free object tracking framework which is efficient and effective. It addresses the long-term standing accuracy misalignment problem of Siamese network based models. Elaborate ablation studies have shown the effectiveness of the whole proposed model and its key components. Without bells and whistles, the proposed method achieves state-of-the-art performance on six tracking benchmarks, with a running speed of 65 FPS.

%The \peng{localization branch} is first proposed to eliminate the gap between classification scores and localization accuracy. Further, reciprocal classification and regression are proposed by re-weighting classification loss with regression quality and regression loss with classification quality. 

\clearpage

\bibliographystyle{named}
\small
\bibliography{ijcai21}

\end{document}

\maketitle

\section{Overview}

This supplementary material includes:

%\noindent(1) The detailed design of the \peng{Localization-Aware} Anchor-Free Head \pjl{in SiamRCR network}.
\begin{itemize}
\item \peng{The comparison of localization prediction methods.}

\item The details of data augmentation in training.

\item \peng{The experiments on OTB-2015 and VOT-2018.}

\item The \pjl{qualitative \ywu{result} comparison \ywu{between} SiamRCR} \ywu{and} other SOTA methods, including SiamRPN++ and SiamFC++.
\end{itemize}

%\section{IoU-Aware Anchor-Free Head}
%The IoU-Aware anchor-free head treats cross-correlation features as inputs. \ywu{As shown in Figure \ref{fig:head}, both} \ywu{the} classification branch and \ywu{the} regression branch contain two 3*3 convolution \ywu{layers} with 256 channels and two ReLU activation \ywu{functions}. After these \ywu{operations}, a 3*3 convolution layer with 1 channel is \ywu{applied in the} classification branch to output \ywu{the} classification confidence map. \ywu{Similarly, two} convolution layers are \ywu{designed} to output \ywu{a} predicted \peng{localization confidence map for the localization branch} and \ywu{bounding box} regression \ywu{results for the regression branch}, respectively.  

\section{Comparison of Localization Prediction}

\begin{figure}[t!]
\centering
\includegraphics[width=0.95\columnwidth]{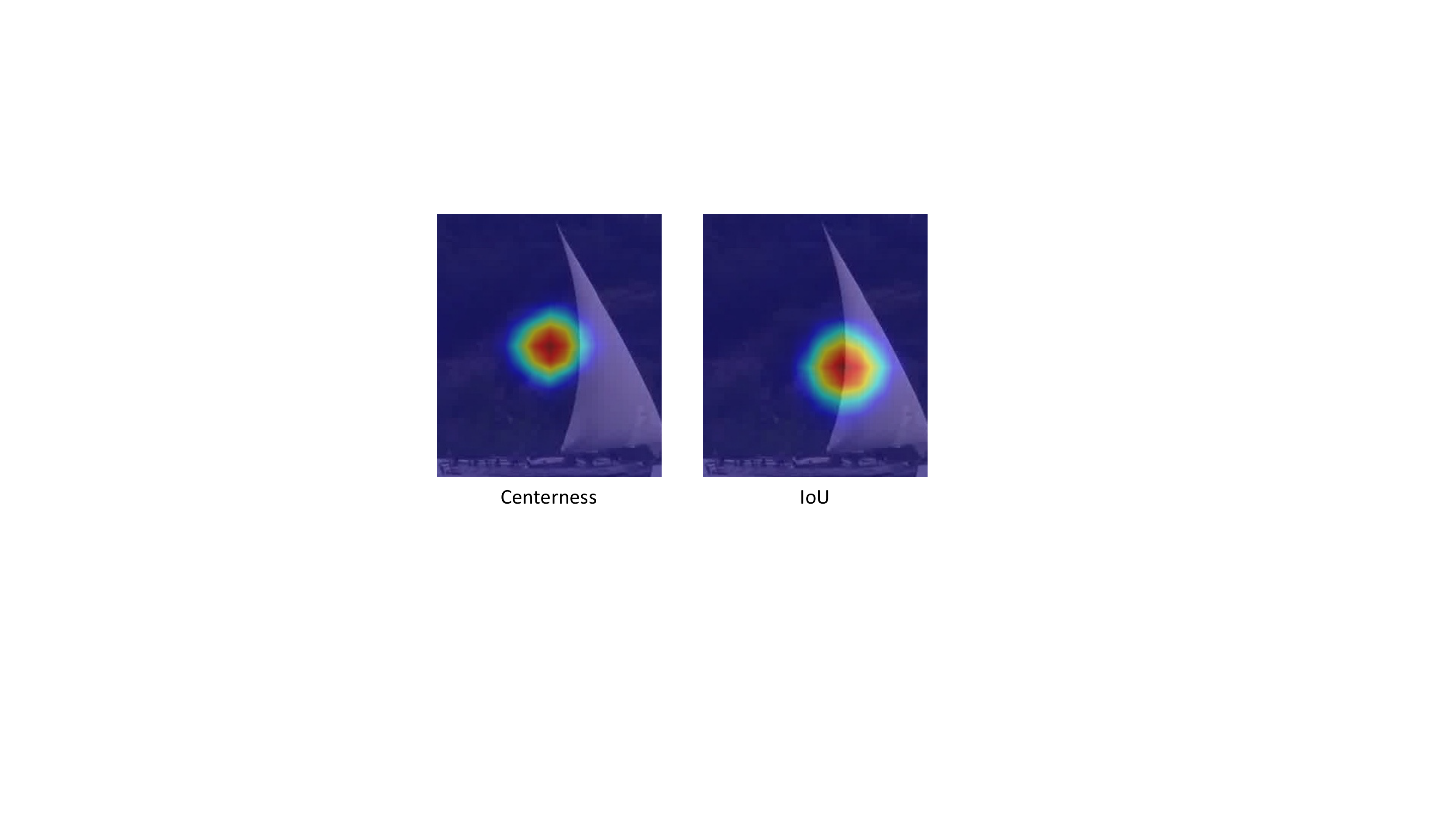}
\caption{The localization \ywu{confidence map} comparison \ywu{between the proposed} localization branch \ywu{and} centerness. For centerness, center location has larger value, which is not suitable for most of ground-truth boxes. Our proposed target is the IoU between the regressed box and the matched ground-truth box, which dynamically changes to match localization accuracy of regressed boxes during training.}
\label{centerness-vs-iou}
\end{figure}

\peng{To demonstrate the \ywunew{superiority} of our \ywunew{dynamically} supervised localization branch compared with centerness, we \ywunew{illustrated} the correlation between the IoU score and the tracking score in the main text. \ywunew{For a more straightforward visual comparison, we} visualize the localization confidence \ywunew{maps 
generated by these two methods on a concrete image sample}. As shown in Figure~\ref{centerness-vs-iou}, \ywunew{The adopted} IoU between the regressed box and the matched ground-truth box in our localization branch could dynamically improve the localization accuracy during training compared with centerness, \ywunew{leading to} a more accurate localization confidence map.}

\iffalse
\begin{figure}[t!]
\centering
\includegraphics[width=\columnwidth]{fig/head.pdf}
\caption{\ywu{Detailed design} of \ywu{the} localization-aware anchor-free Head. It consists of \ywu{a} classification branch (above), \ywu{a} localization branch (middle) and \ywu{a} regression branch (below). }
\label{fig:head}
\end{figure}
\fi

\section{Details of Data Augumentation}
\ywu{In the training of SiamRCR, we collect samples (image pairs) \ywu{following DaSiamRPN}.} The image pairs selected from the same video sequence are regarded as positive samples. The image pairs selected from the static images and different video sequences are regarded as negative samples. We set the ratio of positive \ywu{samples to} negative samples \ywu{to} 3:1. In addition, we \ywu{apply} random translation and \ywu{resizing to the target} images to avoid putting a strong center bias on objects. \ywu{The translation is in the range of 0 to 64 pixels, and the resizing factor is within $[\frac{1}{3}, 3]$.}
%The targets in search images are randomly resized $\frac{1}{3}$ to $3$ times and shifted from 0 to 64 pixels. 

\section{Experiments on OTB-2015 and VOT-2018}

\paragraph{OTB-2015.} OTB-2015 contains 98 sequences and 100 objects. In a quarter sequences of OTB-2015, frames are gray-scale images. Generally, Succ. and Prec. are used as metrics in OTB-2015. Table~\ref{fig:otb} presents the performance of evaluated trackers. Succ. score is the more reasonable metric than Prec. score. since it is normalized by the scale of boxes. Our SiamRCR obtains the best Succ. score, \peng{which demonstrates the effectiveness of SiamRCR.}

\paragraph{VOT-2018.} We report experimental results on VOT-2018 in Table~\ref{fig:vot2018}. We can find that Ocean achieves Top-1 EAO and Top-2 Robustness score on VOT-2018. Our SiamRCR obtains Top-2 EAO and Top-2 Accuracy score on VOT-2018. \peng{Both VOT-2018 and VOT-2019 contains 60 sequences. The difference between them is that VOT-2019 replaces 10 least challenging sequences by \ywunew{another} 10 sequences. VOT-2018 only contains \ywunew{14,687} frames while VOT-2019 contains \ywunew{215,294} frames. Compared with Ocean, the performance of our SiamRCR is higher \ywunew{on all the} benchmarks (including VOT-2019) \ywu{but} VOT-2018, because the VOT-2018 \ywunew{dataset} is relatively simple and easy to overfit. \ywunew{Though we choose to keep its simplicity and purity for more focused presentation and justification, we would like to clarified that SiamRCR is largely extendable and its capability on Robustness can be easily enhanced by introducing existing strategies/solutions, such as \ywu{building an} online module, \ywu{introducing} OA-Conv and motion-aware backbone \ywu{features}. \ywu{Online} trackers such as DiMP50 \ywu{clearly shows that the online modules} can improve robustness, since the model parameters \ywu{can} be updated online during inference. Ocean uses the OA-Conv module to align features with regressed boxes, also alleviating the robustness issue for occlusion, deformation and quick motion phenomenon. \ywu{Partially due to the purity of SiamRCR, its} running speed (65 FPS) is faster than DiMP50 (40 FPS) and Ocean (58 FPS), \ywu{leaving enough room for further extension}}.}

\begin{table}[t]
\small
\centering
\begin{tabular}{ccc}
\toprule[2pt]
          & Succ.$\uparrow$ & Prec.$\uparrow$ \\ \midrule[1pt]
SiamFC    & 0.582 & 0.711 \\
MDNet      & 0.678 & 0.909 \\
GradNet  & 0.639 & 0.861 \\
ATOM      & 0.667 & 0.879 \\
DiMP50    & 0.686 & 0.899 \\
SiamRPN++ & \textbf{\textcolor{blue}{0.696}} & \textbf{\textcolor{red}{0.914}} \\
Ocean     & 0.672 & 0.902 \\
SiamFC++  & 0.683 & 0.896 \\
PG-Net  & 0.691 & 0.892 \\
SiamBAN   & \textbf{\textcolor{blue}{0.696}} & \textbf{\textcolor{blue}{0.910}} \\ \midrule[1pt]
\ywu{SiamRCR (ours)}      & \textbf{\textcolor{red}{0.697}} & 0.900 \\ 
\bottomrule[2pt]
\end{tabular}
\caption{Comparison of tracking results on OTB-2015 benchmark.}
\label{fig:otb}
\end{table}

\begin{table}[t]
\small
\centering
\setlength{\tabcolsep}{0.8mm}{
\begin{tabular}{cccc}
\toprule[2pt]
          & EAO$\uparrow$   & Accuracy$\uparrow$ & Robustness$\downarrow$ \\ \midrule[1pt]
SiamFC    & 0.188 & 0.503    & 0.585      \\
ECO      & 0.280 & 0.484    & 0.276      \\
UPDN      & 0.378 & 0.536    & 0.184      \\
ATOM     & 0.410 & 0.590    & 0.203      \\
SiamRPN++ & 0.414 & \textbf{\textcolor{red}{0.600}}    & 0.234      \\
DiMP50  & 0.440 & 0.597    & \textbf{\textcolor{red}{0.153}}      \\
SiamFC++ & 0.426 & 0.587    & 0.183      \\
SiamBAN  & 0.452 & 0.597    & 0.178      \\
Ocean    & \textbf{\textcolor{red}{0.467}} & 0.598    & \textbf{\textcolor{blue}{0.169}}      \\ \midrule[1pt]
\ywu{SiamRCR (ours)}      & \textbf{\textcolor{blue}{0.457}} & \textbf{\textcolor{blue}{0.599}}    & 0.188     \\ 
\bottomrule[2pt]
\end{tabular}}
\caption{Comparison of tracking results on VOT-2018 benchmark.}
\label{fig:vot2018}

\end{table}

\section{Quantitative Results}
\begin{figure*}
\centering
\includegraphics[width=\textwidth]{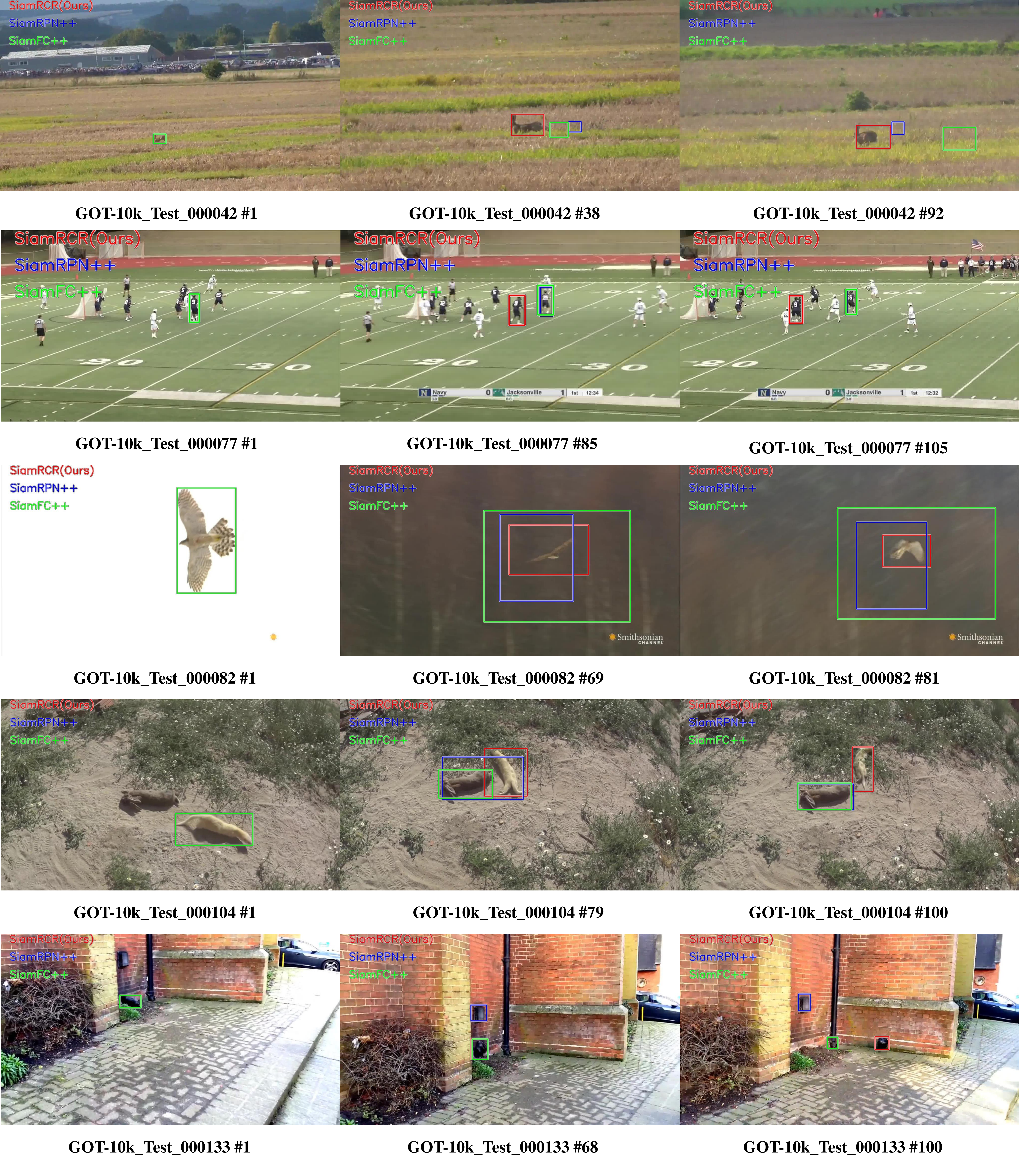}
\caption{Quantitative \ywu{result} comparison \ywu{among} our SiamRCR \ywu{model} (\textbf{\textcolor{red}{red}}), SiamRPN++ (\textbf{\textcolor{blue}{blue}}) and SiamFC++ (\textbf{\textcolor{green}{green}}). \ywu{Note that all the three models share the same initial bounding box in the first frame of each sequence which is shown in green.}}
\label{fig:quantitative}
\end{figure*}

The \ywu{representative} quantitative results of our proposed SiamRCR on \ywu{the test set of} GOT-10k dataset are \ywu{shown in Figure~\ref{fig:quantitative}}. We also present the quantitative results of two representative \ywu{state-of-the-art} trackers: anchor-based SiamRPN++ and anchor-free SiamFC++ \ywu{for a} comparison. \pjl{The} three trackers are initialized with \pjl{the} same ground-truth in the first frame \ywu{of each sequence} which is shown in green in Figure~\ref{fig:quantitative}.

\ywu{Figure~\ref{fig:quantitative} demonstrates} that SiamRPN++ and SiamFC++ may fail to \pjl{track} \ywu{the} targets in cases of fast motion, scale variation and occlusion. In sequence 42, SiamRPN++ and SiamFC++ drift from the moving animal in frame 38. Our proposed \pjl{SiamRCR} can locate \ywu{the target} \pjl{accurately} with more \pjl{reasonable} localization confidence \pjl{\ywu{thanks to the reciprocal links which integrate} the classification confidence score and predicted \peng{localization} score}. In \ywu{sequences} 77 and 104, there are distractors \ywu{which have similar appearances as the} targets in the scene. \ywu{Both SiamRPN++ and SiamFC++ get confused and switch to the distractors while our SiamRCR keeps tracking the right targets.} %Our SiamRCR tracks targets according to not only classification confidence but also localization confidence. In the above-mentioned cases, it distinguishes two object of same class.
%\pjl{Differently, by building reciprocal links between classification and regression branches, our proposed SiamRCR is able to distinguish the two object of same class in these cases.}
In sequence 82, SiamRCR \ywu{can quickly adapt to the great} scale \ywu{variations} of the flying \pjl{eagle} \ywu{despite the motion blur and low foreground-background contrast, while the other two competitors fail to do that}. \pjl{In sequence 133, the target is a \ywu{flying} black \ywu{plastic} bag. SiamRPN++ and SiamFC++ \ywu{get distracted by some other} black \ywu{object or object part} in frame 68 and frame 100, respectively, while our SiamRCR \ywu{successfully ignores} the interference and keeps tracking the initial \ywu{target}.}

\pjl{More complete and clear visualization \ywu{on the} tracking \ywu{result} comparison is displayed in the video attachments.}

\clearpage

%The IoU branch is first proposed to eliminate the gap between classification scores and localization accuracy. Further, reciprocal classification and regression are proposed by re-weighting classification loss with regression quality and regression loss with classification quality. 

%\bibliographystyle{named}
%\bibliography{ijcai21}